\documentclass[10pt,twocolumn,letterpaper]{article}

\usepackage{wacv}              

\usepackage{graphicx}
\usepackage{amsmath}
\usepackage{amssymb}
\usepackage{booktabs}

\usepackage{amssymb}
\usepackage{graphicx}
\usepackage{amsmath}
\usepackage{amssymb}
\usepackage{booktabs}
\usepackage{tabu}
\usepackage[T1]{fontenc}
\usepackage{xcolor}
\usepackage{array}
\usepackage{epsfig}
\usepackage{epstopdf}
\usepackage{booktabs}
\usepackage{algorithm}
\usepackage{algpseudocode}
\usepackage{lipsum}
\usepackage{multirow}
\usepackage{textcomp}
\usepackage{gensymb}
\usepackage{pifont}

\usepackage{tabularx}
\usepackage{adjustbox}
\usepackage{array}
\usepackage{multirow}
\usepackage{url}
\usepackage{booktabs}
\usepackage{array}
\usepackage{lipsum}
\usepackage{makecell} 
\usepackage{xcolor}
\usepackage{caption, subcaption}

\newcommand\blfootnote[1]{%
\begingroup 
\renewcommand\thefootnote{}\footnote{#1}%
\addtocounter{footnote}{-1}%
\endgroup 
}
\usepackage[pagebackref,breaklinks,colorlinks]{hyperref}

\usepackage[capitalize]{cleveref}
\crefname{section}{Sec.}{Secs.}
\Crefname{section}{Section}{Sections}
\Crefname{table}{Table}{Tables}
\crefname{table}{Tab.}{Tabs.}

\def\wacvPaperID{1207} 
\def\confName{WACV}
\def\confYear{2024}

\begin{document}

\title{ShARc: Shape and Appearance Recognition for Person Identification In-the-wild}

\author{Haidong Zhu\ \ \ \ \ \ \ Wanrong Zheng\ \ \ \ \ \ \ Zhaoheng Zheng\ \ \ \ \ \ \ Ram Nevatia\\
University of Southern California\\
{\tt\small {\{haidongz,wanrongz,zhaoheng.zheng,nevatia\}@usc.edu}}
}
\maketitle

\begin{abstract}
Identifying individuals in unconstrained video settings is a valuable yet challenging task in biometric analysis due to variations in appearances, environments, degradations, and occlusions. In this paper, we present ShARc, a multimodal approach for video-based person identification in uncontrolled environments that emphasizes 3-D body shape, pose, and appearance. We introduce two encoders: a Pose and Shape Encoder (PSE) and an Aggregated Appearance Encoder (AAE). PSE encodes the body shape via binarized silhouettes, skeleton motions, and 3-D body shape, while AAE provides two levels of temporal appearance feature aggregation: attention-based feature aggregation and averaging aggregation. For attention-based feature aggregation, we employ spatial and temporal attention to focus on key areas for person distinction. For averaging aggregation, we introduce a novel flattening layer after averaging to extract more distinguishable information and reduce overfitting of attention. We utilize centroid feature averaging for gallery registration. We demonstrate significant improvements over existing state-of-the-art methods on public datasets, including CCVID, MEVID, and BRIAR.
\blfootnote{This research is based upon work supported in part by the Office of the Director of National Intelligence (ODNI), Intelligence Advanced Research Projects Activity (IARPA), via [2022-21102100007]. The views and conclusions contained herein are those of the authors and should not be interpreted as necessarily representing the official policies, either expressed or implied, of ODNI, IARPA, or the U.S. Government. The U.S. Government is authorized to reproduce and distribute reprints for governmental purposes notwithstanding any copyright annotation therein.}
\end{abstract}
\section{Introduction}

Recognizing individuals in-the-wild \cite{nalty2022brief} is a challenging yet valuable task for determining a person's identity from images or videos, playing a crucial role in many applications. Since face images may be unreliable or unavailable for individuals at a distance or from specific viewpoints, recognizing individuals via body images or videos becomes increasingly important. In this paper, we focus on video-level appearance and body shapes to develop a robust identification system suitable for various distances and camera viewpoints, utilizing multiple videos as gallery samples. We specifically address different clothing and activities in generalized scenarios by comparing and combining shape and appearance-based methods for identification.

\begin{figure}[t]
    \centering
    \resizebox{0.92\linewidth}{!}
    {
    \begin{tabular}{p{1.7cm}p{1.7cm}<{\centering}p{1.7cm}<{\centering}p{1.7cm}<{\centering}} 
     \centering
     \begin{subfigure}[b]{0.8\linewidth}
         \centering
         \includegraphics[width=\textwidth]{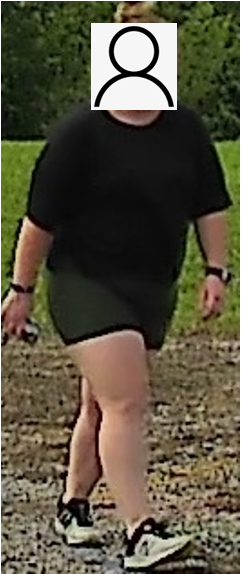}
     \end{subfigure} &
     \begin{subfigure}[b]{0.8\linewidth}
         \centering
         \includegraphics[width=\textwidth]{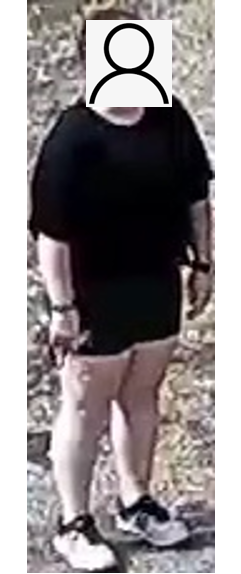}
     \end{subfigure} &
     \begin{subfigure}[b]{0.8\linewidth}
         \centering
         \includegraphics[width=\textwidth]{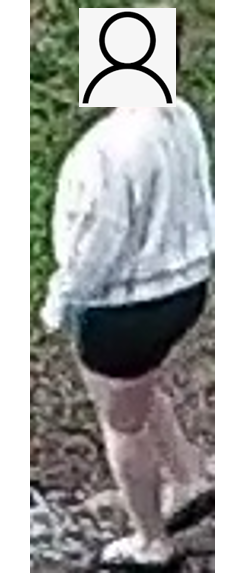}
     \end{subfigure} &
     \begin{subfigure}[b]{0.8\linewidth}
         \centering
         \includegraphics[width=\textwidth]{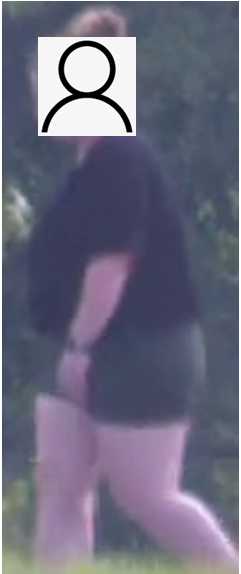}
     \end{subfigure}
     \\
    \toprule
    \small{Gallery Frame} & \small{Standing Videos} & \small{Different Clothing} & \small{Turbulence \& Occlusion}\\
    \midrule
    \small{Gait} &  & \checkmark& \checkmark\kern-1.1ex\raisebox{.7ex}{\rotatebox[origin=c]{125}{--}} \\
    \small{Body shape} & \checkmark\kern-1.1ex\raisebox{.7ex}{\rotatebox[origin=c]{125}{--}} & \checkmark\kern-1.1ex\raisebox{.7ex}{\rotatebox[origin=c]{125}{--}} & \checkmark \\
    \small{Appearance} & \checkmark &\checkmark\kern-1.1ex\raisebox{.7ex}{\rotatebox[origin=c]{125}{--}} & \checkmark\kern-1.1ex\raisebox{.7ex}{\rotatebox[origin=c]{125}{--}} \\
    \midrule
    \small{Ours} & \checkmark & \checkmark & \checkmark\\
    \bottomrule
    \end{tabular}
    }
    \smallskip
    \caption{To identify a person, gait is unreliable in stationary videos, and appearance alters when subjects wear different clothing. The imprecise reconstruction of 3-D body shapes results in unstable predictions while the human prior assists in occlusions.}
    \label{fig:example}
\end{figure}

To identify individuals from their body, research primarily focuses on appearance \cite{wieczorek2021unreasonable,wang2021pyramid} and gait \cite{chao2019gaitset,fan2020gaitpart,lin2021gaitgl,zheng2022gait,zhu2023gait,zhu2023gaitref}. Unlike facial features, which are relatively constant \cite{deng2019arcface,kim2022adaface,duan2019uniformface}, body appearance can vary significantly due to changes in clothing, environment, and occlusions \cite{davila2023mevid}, as depicted in Figure~\ref{fig:example}. Gait analysis captures an individual's walking pattern and is less affected by environmental changes or clothing. However, it requires a walking sequence that may not always be available. Additionally, varying environmental conditions pose challenges in feature registration and matching, making the prediction of human identity more sensitive to noisy samples in gallery videos.

We introduce ShARc, a method based on \textbf{SH}ape and \textbf{A}ppearance \textbf{R}e\textbf{C}ognition. Specifically, we employ a Pose and Shape Encoder (PSE) and an Aggregated Appearance Encoder (AAE) to project the input video into their corresponding embedding spaces. Leveraging body shapes with shape and motion representations \cite{zhu2023gait}, ShARc enables identification in diverse scenarios; a robust body prior \cite{loper2015smpl} offers guidance under occlusion or variations in clothing. Alongside this, we introduce multi-level appearance features for both video-level and frame-level analysis. Importantly, these techniques show commendable performance even before combining with body shapes.

To extract the shape of a person in a sequence, we disentangle motion and poses by extracting skeletons, 3-D body shapes, and silhouettes from tracklets with our Pose and Shape Encoder (PSE). We utilize silhouettes and 3-D body shapes to represent individual frame shape patterns in 2-D and 3-D space, while employing sequential skeletons to represent motions. For the two different shape modalities, we first extract their frame-wise features and then combine them frame-by-frame using an attention mechanism for body shape feature extraction. Subsequently, we concatenate the pooled features with pose features encoded from skeletons for the final shape representation.

Parallel to body shape extraction, we also use an Aggregated Appearance Encoder (AAE) to extract features from appearances, preserving identification information from raw images. We obtain both frame-wise and video-level features and integrate them for dual-level understanding. For frame-level extraction, we introduce a novel flattening layer after averaging to extract more distinguishable information and reduce overfitting. At the video level, we employ spatial and temporal attention, as per \cite{wang2021pyramid}, to focus on key areas for person distinction. This allows the model to concentrate on unique patterns in both frame and sequence.

After obtaining both shape and appearance features, we employ centroid feature averaging for gallery registration, using the mean features of the same ID rather than comparing to each gallery separately. This helps to mitigate variances in gallery examples with different clothing. We validate our approach on public datasets like CCVID \cite{gu2022clothes}, MEVID \cite{davila2023mevid}, and the recently-released BRIAR \cite{cornett2023expanding}, showing state-of-the-art performance on all of them.

In summary, our contributions are as follows: 1) We introduce ShARc, a multimodal method for person identification in-the-wild using video samples, focusing on both shape and appearance; 2) We unveil a novel Pose and Shape Encoder (PSE) that captures dynamic motion and body shape features for more robust shape-based identification; 3) We deploy an Aggregated Appearance Encoder (AAE) that incorporates both frame-level and video-level features.
\section{Related Work}


\textbf{Person Identification Based on Body Appearance} is a critical task in computer vision that focuses on identifying and matching individuals across different camera views or separate instances \cite{zheng2015scalable,li2014deepreid}. Unlike face recognition \cite{deng2019arcface,kim2022adaface,duan2019uniformface}, body appearance-based re-identification \cite{zheng2017person,wieczorek2021unreasonable,hong2021fine,jin2022cloth} requires less subject cooperation and is achievable in diverse environments. With deep learning advancements, researchers focus on various ways to extract maximally useful information from single-frame inputs. Approaches include part-based methods \cite{zhu2022pass,cho2022part,li2021diverse,he2021partial,zhao2017deeply,sun2018beyond} and attention \cite{ji2020attention,gao2022semantic} to address occlusions and others.

Besides single-frame person identification, recent research has explored video-level re-identification methods \cite{hou2020temporal,gu2022clothes,wang2021pyramid} by introducing more frames and reducing poor-quality frame impact for enhanced temporal robustness. Since the model only needs to output one person ID prediction for multiple frames, researchers either use temporal pooling \cite{li2018diversity,liu2017video,zheng2016mars,gao2018revisiting} or recurrent networks \cite{dai2018video,mclaughlin2016recurrent,zhou2017see} for fusing frames across timestamps. Recently, attention mechanisms \cite{hou2020temporal,wang2021pyramid,si2018dual,subramaniam2019co,xu2017jointly,fu2019sta,liu2019spatially,li2019global} have been utilized for aggregating useful information from temporal and spatial dimensions for identification. However, most methods focus on videos with consistent clothing, limiting model generalizability. Researchers are now emphasizing videos with different outfits and environments \cite{gu2022clothes,davila2023mevid}, making identification tasks more applicable for real-life scenarios.

\begin{figure*}[t]
    \centering
    \includegraphics[width=\linewidth]{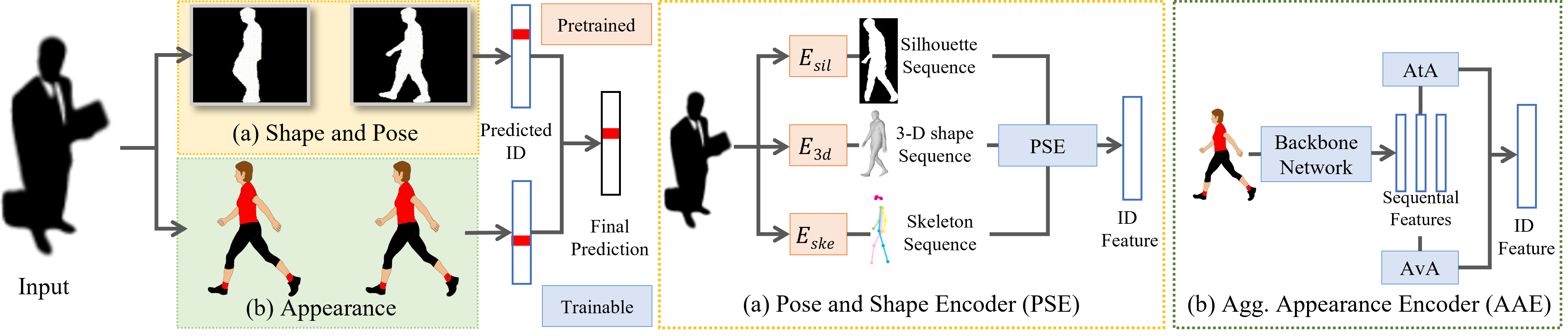}
    \caption{Our proposed method includes two sub modules: (a) a shape-based recognition system, PSE, which extracts the silhouette, 3-D body shape and skeletons sequences and fuses them for person recognition, and (b) an appearance-based recognition system, AAE, which takes both outputs from attention-based aggregation (AgA) and averaging aggregation (AvA) as input for identification.}
    \label{fig:pipeline}
\end{figure*}

\textbf{Gait Recognition} focuses on identifying a person based on their walking patterns. Compared with appearance-based recognition methods, gait patterns, usually captured via binarized silhouettes \cite{yu2006framework,takemura2018multi} describing body shape contours, reduce the negative impact of clothing changes for identification but introduce different appearance variations with body contours. Due to the lack of RGB patterns, it is challenging to infer body information directly from silhouettes. To address this, some researchers \cite{fan2020gaitpart,lin2021gaitgl} focus on part-based recognition, while others \cite{hou2020gln,chao2019gaitset,fan2022opengait} extract framewise consistencies for identification.

Due to the limited information in silhouettes, recent research \cite{an2020performance,zhu2023gait,teepe2021gaitgraph,shen2022lidar,guo2022multi} focuses on external modalities to assist silhouettes for identification. GaitGraph~\cite{teepe2021gaitgraph}, GaitMix \cite{zhu2023gaitref} and GaitRef \cite{zhu2023gaitref} apply or refine HRNet~\cite{wang2020deep} for joint detection and uses the generated pose sequence for identification. Gait3D \cite{zheng2022gait}, GaitHBS~\cite{zhu2023gait}, and ModelGait~\cite{li2020end} focus on extracting or using body shapes alongside silhouettes for gait recognition, intending to provide more information for part separation. LiDARGait~\cite{shen2022lidar} employs point clouds instead of silhouettes for body shape description. Some researchers \cite{liang2022gaitedge,guo2022multi} also integrate RGB images with silhouettes for gait understanding. Since these methods still focus on gait representation, 
they can only apply to walking sequences for identification. Our proposed PSE combines pose with 3-D body shape for identification, inherently removing the requirement for walking sequences.

\section{Methodology}

Given a video with sequential frames $V = \{f_i\}_n$ containing $n$ frames of the person, ShARc decomposes it into two branches: the body shape $\{b_i\}$ and the RGB appearance of the frames $\{a_i\}$ that exhibit the most distinguishable patterns, as illustrated in Figure~\ref{fig:pipeline}. By estimating their independent similarities $S_{shape}(V)$ and $S_{app}(V)$ compared with gallery candidates, ShARc combines the two scores together using weighted average for the final similarity $S(V)$.

\subsection{Shape-based Person Recognition}\label{sec:sha}
For shape-based person recognition, we mitigate the influence of appearance by focusing on alternative representations, such as 3-D human body shape and silhouettes, to emphasize the individual's body shape, as well as skeletons to capture motion in pose. Although gait recognition is useful when walking segments are available, it offers limited distinguishable information in stationary videos when the person is not walking. Unlike existing gait recognition methods \cite{chao2019gaitset, lin2021gaitgl, zhu2023gait, teepe2021gaitgraph}, our shape-based approach compensates for the absence of gait by leveraging extra body shape priors. We first extract the corresponding modalities utilized in our model, which include 3-D body shapes, skeletons, and silhouettes, and then fuse them as the final representation.

\textbf{Shape and motion extraction.} For shape-based person recognition, we focus on two crucial representations for distinguishing individuals: body shape $P_i$ and motion $M_i$. Body shape encompasses specific actions or shapes a person may exhibit, while motion refers to the temporal information, representing a more specific case. If both shape and motion exist in all sequences, the task can be regarded as gait recognition.
For body shape extraction, we focus on two distinct modalities: silhouettes and 3-D body shapes. Silhouettes represent the 2-D human boundary in each frame, while 3-D body shape reconstruction remains invariant to viewpoints by reconstructing the person's 3-D shape. The combination of silhouettes and body shapes allows for the preservation of both general shape and frame-wise detailed reconstruction of the individual.

In addition to body shape, we incorporate skeletons to understand motions, as motions represent the specific movement patterns of a person. Unlike gait recognition tasks \cite{chao2019gaitset}, which use binarized silhouettes as input, skeletons can provide temporal understanding without the biases of body shape. Furthermore, by separating body shapes from motion analysis, the network for pose extraction can better focus on the general shape, aiding temporal understanding and helping the model to maximize the utilization of potential information in the sequence.

For the three modalties described above, we employ three extractors, $E_{sil}(\cdot)$, $E_{3d}(\cdot)$ and $E_{ske}(\cdot)$, to encode the corresponding representations of these three modalities for each frame $i$ following
\begin{equation}
    P_i = E_{sil}(f_i) + E_{3d}(f_i);\ \ M_i = E_{ske}(f_i)
\end{equation}
and extract the corresponding body shape $P_i$ and motion $M_i$ inputs for further processing. For silhouette input, we concatenate the silhouette and the cropped RGB images using silhouette as masks, as our input, since this can provide more separation of the human part in the body shape. Since these modals requires heavy training to ensure a stable performance, we use  pretrained networks to extract these representations, which we discuss in Section~\ref{sec:expd}.

\begin{figure}[t]
    \centering
    \includegraphics[width=\linewidth]{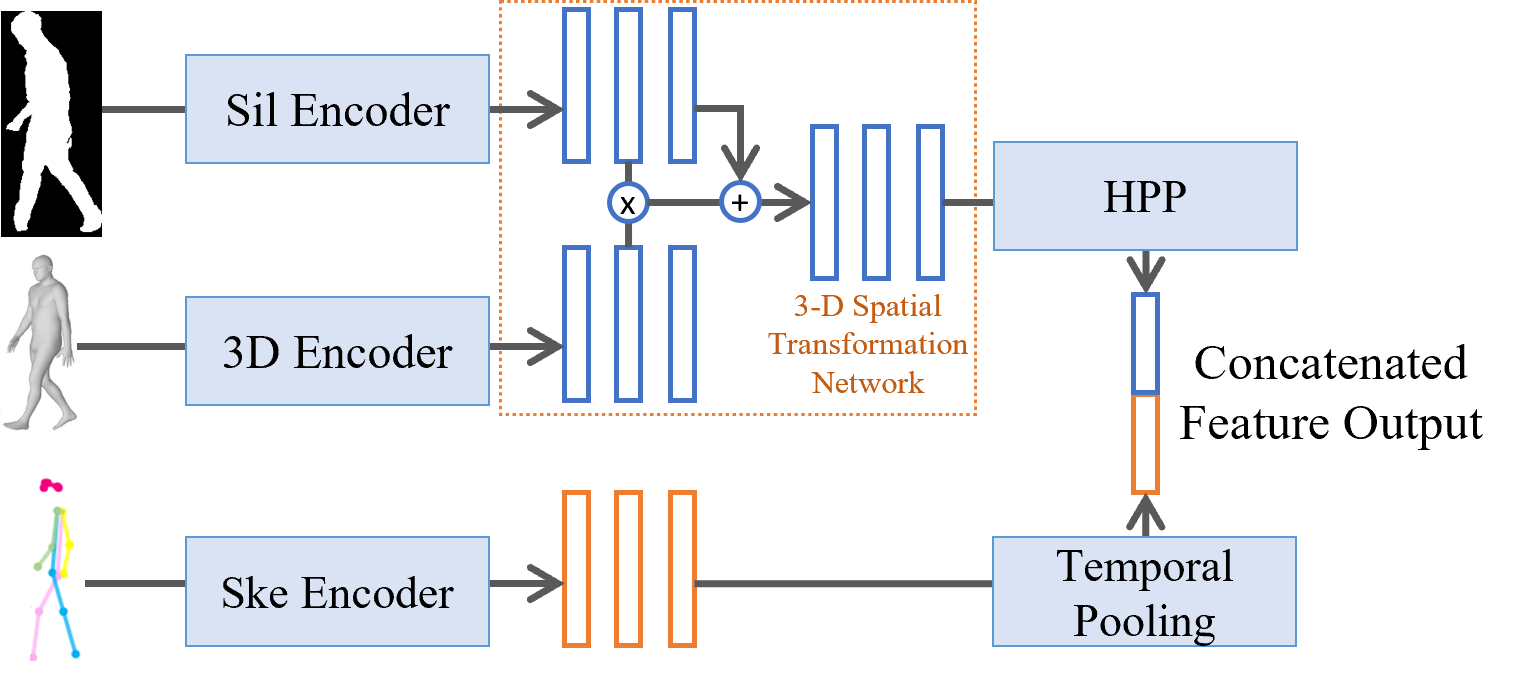}
    \caption{Architecture of PSE for combining body shape and motion information for shape-based identification.}
    \label{fig:gmpp}
\end{figure}

\textbf{Multimodal Fusion.} With these three modalities, we introduce PSE for combining framewise body shape features $P_i$ along with motion pattern $M_i$, as illustrated in Figure~\ref{fig:gmpp}. For feature representation of silhouettes $Feat_{sil}$ and 3-D body shapes $Feat_{3d}$, we use corresponding encoders $F_{pose}$ to project $E_{sil}(f_i)$ and $E_{3d}(f_i)$ into their embedding space. We then apply the 3-D spatial transformation network \cite{zheng2022gait} with skip connection and implement horizontal pyramid pooling $HPP$ \cite{chao2019gaitset} with $B$ bins after the encoder output for each frame following
\begin{align}
\begin{split}
I_{sil},\ I_{3d}&= F_{pose}(E_{sil}(f_i),\ E_{3d}(f_i))\\
I_{pose} &= (I_{sil} \cdot I_{3d}) + I_{sil}\\
I_{pose} &= HPP(I_{pose})
\end{split}
\end{align}
where $I_k$ represents the feature for the modality $k$. For motion representation, we utilize a motion encoder $F_{motion}$ to extract multi-level spatial and temporal skeleton information and use average pooling along the temporal dimension for the generated feature of the last layer. Then, we concatenate the skeleton feature, after temporal pooling, along with the pose representation as an additional new bin in the matching process, making the concatenated $(B + 1) \times C$ feature map our final output for shape representation:
\begin{align}
\begin{split}
I_{motion}&= AvgPooling(F_{motion}(E_{ske}(f_i)))\\
I_{shape} &= [I_{pose}, I_{motion}]
\end{split}
\end{align}
where $[\cdot,\cdot]$ represents feature concatenation.

\subsection{Appearance-based Person Recognition}\label{sec:app}
Compared to shape-based methods, which depend on the accuracy of body shape and contours, appearance provides richer and lossless RGB information for distinguishing individuals. We implement both attention-based and averaging appearance aggregation for identification. As people may wear different clothing and be in varying environmental conditions, we 
incorporate temporal and spatial information with attention-based appearance aggregation to focus on the relevant parts for differentiation between nearby frames. Moreover, to avoid overfitting on specific body parts or frames, we also employ video-level averaging aggregation to equally utilize spatial and temporal features.

\begin{figure}[t]
    \centering
    \includegraphics[width=\linewidth]{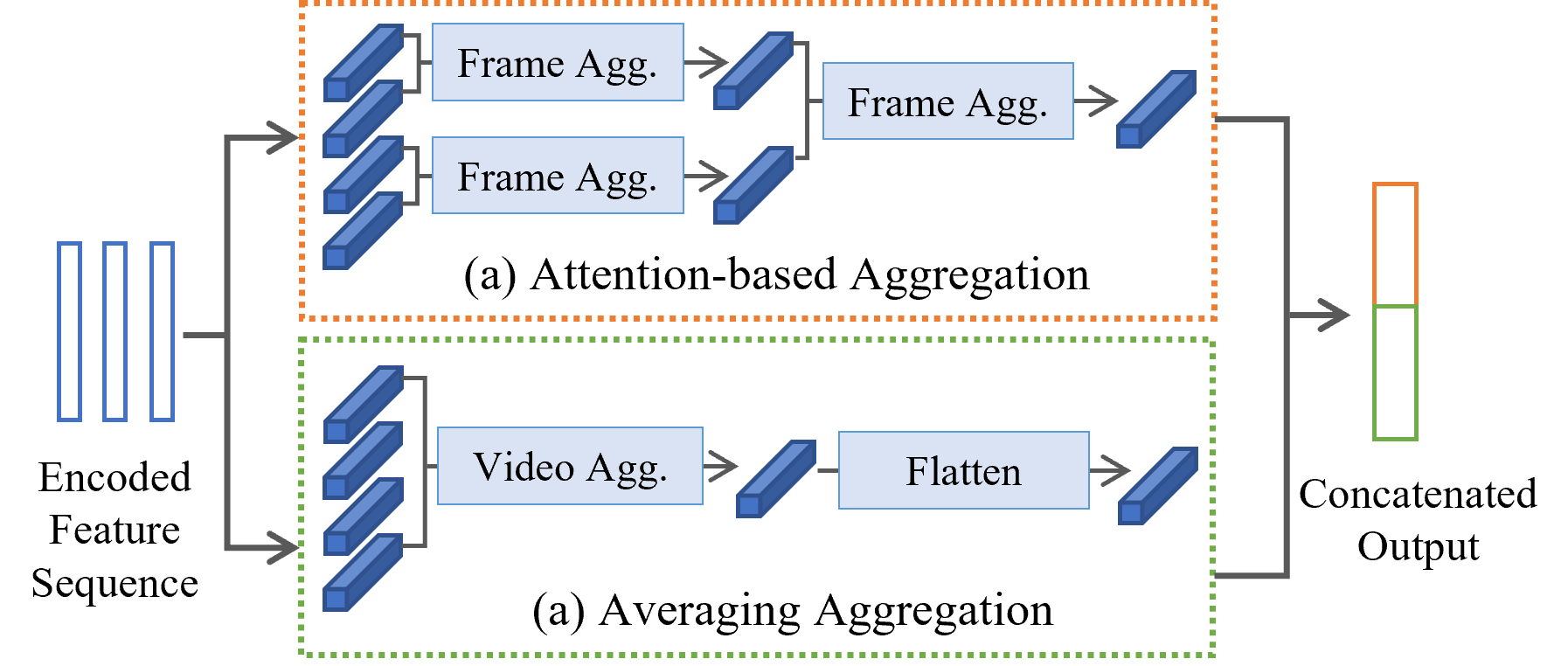}
    \caption{Architecture of the AAE with an example of sequence length $n=4$. AAE aggregate the video frames in two ways: 1) attention-based aggregation, which mines the connection between nearby frames with attention, and 2) averaging aggregation, which takes all the frames together equally.}
    \label{fig:appearance}
\end{figure}

\textbf{Attention-based Aggregation.} 
For attention-based aggregation, we follow Figure~\ref{fig:appearance} (a) for building spatial and temporal attention (STA) for the features extracted from the backbone network, encoding each frame $F_i$ to their corresponding features $A_i$. We follow \cite{wang2021pyramid} to combine the features of two frames using a 3-level pyramid following
\begin{align}
\begin{split}
A_t^{l+1} = SA(A_t^l) + SA(A_{t+1}^l) + TA(A_t^l, A_{t+1}^l)
\end{split}
\end{align}
where $l$ is the current layer in the pyramid, and $t$ is the temporal stamp for the current frame. $TA$ and $SA$ are two attention generation layers following \cite{wang2021pyramid}. For each layer of the pyramid, we reduce the number of available appearance features to half the size of its previous layer, until we get the output feature representation in the last layer. This means the network, as an example, can handle at most 8 frames for the final feature $A_{attn}(V)$ with a three layers of pyramid. It is important to note that if attention-based aggregation is not combined with averaging aggregation and its backbone feature encoder not shared, it is degraded to the existing method PSTA \cite{wang2021pyramid} encoder.

\textbf{Averaging Aggregation.} 
As attention mechanism may create overfitting when there is shift between training and testing domain, we add averaging aggregation, as illustrated in Figure~\ref{fig:appearance} (b), for global representation extraction. Video-level appearance focuses on finding the corresponding features of each frame and treating all the frames equally.
After extracting the framewise appearance feature $A_i$, we average the features of all the frames in the same video following
\begin{align}
\begin{split}
A_{avg} &= \frac{1}{n}\sum_{i=1}^n A_{i}
\end{split}
\end{align}

We then use Gamma Correction $\gamma$ in the range of $[0, 1]$ to flatten the features as a feature flatten layer following
\begin{align}
\begin{split}
A_{avg} &= sgn(A_{avg}) \cdot ||A_{avg}||^\gamma
\end{split}
\end{align}
where $sgn(\cdot)$ is the sign function operated on channel-wise elements. Since the videos include multiple frames that may capture the person from different aspects, some of the specific representative features of this person may not be captured in all the frames. With $\gamma < 1$, the new feature are different from the old one in cases. When the feature value is close to the zero point (0), flattening layer makes the original value more distinguishable by increasing its absolute value. In addition, the flattening layer can also reduce the maximum value and avoid overfitting with feature values far from 0, making the network focus on more patterns instead of on just a few of them for making predictions.

\subsection{Registration and Fusion}\label{sec:fusion}
For person identification in the wild, it is essential to handle videos of individuals with varying clothing conditions, as gallery videos also exhibit differences in clothing, leading to variances in appearance. To address this issue, we follow \cite{wieczorek2021unreasonable} and construct a centroid representation for registering gallery examples. Assuming we have $k\times c$ features with a same ID, we average the $k$ features and use the $1\times c$ feature for representing this ID in the gallery. The averaging operation helps to mitigate the biases arising from different clothing, as clothing across videos are assumed to be randomly distributed, while the appearance remains consistent.

Since shape and appearance are distinct modalities, we compute the features independently for each and match them with their corresponding modalities in the gallery candidates to obtain two matching scores, $S_{shape}(V)$ and $S_{app}(V)$. We then use a weighted average function to combine these two scores following
\begin{equation}
S(V) = \alpha S_{shape}(V) + (1 - \alpha) S_{app}(V),
\end{equation}
where $S(V)$ is the final similarity score, $S_{shape}(V)$ and $S_{app}(V)$ are the shape-based and appearance-based similarity scores, respectively, and $\alpha$ is a weight parameter that balances the contributions of the two modalities. By adjusting $\alpha$, we can find the optimal combination that leads to the best overall identification performance. Based on our ablation results in Section~\ref{sec:result}, we set $\alpha$ to 0.1 in our experiment.

\subsection{Objectives}\label{sec:objective}
Considering that ShARc is a model for video-based identification, we train shape and appearance models separately, using end-to-end training for each. For the shape-based recognition model, PSE, we follow \cite{zheng2022gait} and combine triplet loss $\mathcal{L}_{triplet}$ \cite{schroff2015facenet} with a margin of 0.2, along with cross-entropy loss $\mathcal{L}_{CE}$ as follows:
\begin{equation}
\mathcal{L}_{shape} = 0.1\ \mathcal{L}_{triplet} + \mathcal{L}_{CE}
\end{equation}
For the appearance model, we apply four losses following \cite{wieczorek2021unreasonable}, which combines a Triplet loss $\mathcal{L}_{triplet}$ \cite{schroff2015facenet} with 0.3 as margin, a Center Loss $\mathcal{L}_{cen}$ \cite{wen2016discriminative}, a Cross Entropy loss $\mathcal{L}_{CE}$, and a Centroid Triplet Loss $\mathcal{L}_{CTL}$ \cite{wieczorek2021unreasonable}, as follows:
\begin{equation}
\mathcal{L}_{app} = \mathcal{L}_{triplet} + \mathcal{L}_{CE} + \mathcal{L}_{cen} + 5e^{-4}\ \mathcal{L}_{CTL}
\end{equation}

\section{Experiments and Results}

\subsection{Experimental Details}\label{sec:expd}

\textbf{Datasets.} In our experiment, we primarily compare our method with other state-of-the-art methods on three challenging, public, video-based datasets: CCVID \cite{gu2022clothes}, MEVID \cite{davila2023mevid}, and BRIAR \cite{cornett2023expanding}. We include the statistics for these three datasets in Table~\ref{table:dataset}. CCVID \cite{gu2022clothes} and MEVID \cite{davila2023mevid} are recent datasets featuring the same and different clothes and include more than one outfit for each identity, with 226 and 158 identities, respectively. Unlike CCVID, which has only one viewpoint from the same location, MEVID includes 33 viewpoints and multiple scales of images from 33 different settings. BRIAR is a large, in-the-wild person identification dataset with varying distances, conditions, activities, and outfits for identification.

Compared to CCVID and MEVID, BRIAR \cite{cornett2023expanding} encompasses more variations of distances, viewpoints, and candidate IDs, which models the person identification problem in the wild. In addition, BRIAR has more distractor IDs in the gallery for the open-set problem evaluation, as well as featuring more images from elevated cameras and UAVs, introducing greater difficulty for final template matching. Since the BRIAR dataset is continuously expanding, we use the version including both BGC1 and 2 following \cite{cornett2023expanding}, which is an extended version compared to \cite{guo2022multi}.

\begin{table}
\begin{center}
\resizebox{0.86\columnwidth}{!}
{
\begin{tabu}{p{2.4cm}<{\centering}p{1.2cm}<{\centering}p{1.5cm}<{\centering}p{1.5cm}<{\centering}p{1.5cm}<{\centering}}
\toprule
Dataset & Split & \#frames & \#identities & \#tracklets\\
 \midrule
\multirow{3}{*}{BRIAR}& train  & 4,366,198 & 407 & 37,466 \\
& query &189,819 & 192 & 886\\
& gallery & 2,326,111 & 544 & 4,379 \\
\midrule
\multirow{3}{*}{CCVID}& train &116,799  &75 & 948\\
& query & 118,613&151 &  834\\
& gallery &112,421 &151  & 1,074\\
\midrule
\multirow{3}{*}{MEVID}& train & 3,609,156 & 104 & 6,338\\
& query & 205,044 & 52 & 316\\
& gallery & 981,207 & 54 & 1,438\\
\bottomrule
\end{tabu}}
\end{center}
\caption{Statistics for the three datasets in our experiment.}
\label{table:dataset}
\end{table}

\textbf{Implementation Details.} We first discuss the detailed architecture used for shape and appearance-based networks, followed by the training and inference details.

\textit{Shape-based Modalities Extraction.} For shape-based recognition, our model requires three different inputs: silhouettes, 3-D body shapes, and skeletons. For silhouette extraction $E_{sil}(\cdot)$, we use DeepLab-v3 \cite{chen2017rethinking} with ResNet-101 \cite{he2016deep} pretrained on the Pascal VOC dataset as the backbone to identify the pixels predicted as the 'person' category for silhouettes. For the 3-D human body shape extraction $E_{3d}(\cdot)$, we use ROMP \cite{sun2021monocular} pretrained on Human3.6M \cite{ionescu2013human3} and MPI-INF-3DHP \cite{mehta2017monocular} to extract three vectors: a 3-D camera parameter, a 10-D vector body shape, and a 72-D vector representing the rotation of the joints. These three vectors form an 85-D SMPL \cite{loper2015smpl,zhu2022open} representation for each frame. Since there is only one person in each frame sequence, we use the first SMPL body shape predicted by ROMP as our body shape representation. For skeletons $E_{ske}(\cdot)$, we follow \cite{teepe2021gaitgraph} and use HRNet \cite{sun2019deep} with architecture `pose\_hrnet\_w32' and 384$\times$288 as input size, which is pretrained on the MS COCO dataset \cite{lin2014microsoft} for 2-D pose estimation as the skeleton representation.

With different input modalities available for shape-based modal extraction, we use ResNet-9 \cite{fan2022opengait} as the gait encoder, a 4-layer MLP \cite{zheng2022gait} for 3-D body shape encoding, and MS-G3D \cite{liu2020disentangling} for skeleton encoding. All these three models are trained together with PSE end-to-end with the shape-based recognition model.

\begin{table}[tb]
\centering
\def\lw{1.3}
\def\lk{1.4}
\def\ls{0.0}
\resizebox{\linewidth}{!}
{
\begin{tabular}{p{2.95cm}p{\ls cm}p{\lw cm}<{\centering}p{\lw cm}<{\centering}p{\ls cm}p{\lw cm}<{\centering}p{\lw cm}<{\centering}p{\ls cm}p{\lw cm}<{\centering}p{\lw cm}<{\centering}}
\toprule

\multirow{2}{*}{Method} && \multicolumn{2}{c}{All Activities} && \multicolumn{2}{c}{Walking Sequences} && \multicolumn{2}{c}{Stationary Sequences} \\

 \cline{3-4} \cline{6-7} \cline{9-10}  \\ [-8pt]
            && Rank 1     & Rank 20  && Rank 1    & Rank 20 && Rank 1   &  Rank 20  \\
\midrule
GaitSet \cite{chao2019gaitset} && 15.3 & 40.5 && 27.7 & 64.5 && 7.3 & 24.9 \\
GaitPart \cite{fan2020gaitpart} && 14.1 & 41.7 && 25.7 & 67.8 && 6.6 & 24.8 \\
GaitGL \cite{lin2021gaitgl} && 15.6 & 45.1 && 28.0 & 67.2 && 7.5 & 30.8 \\
GaitMix \cite{zhu2023gaitref} && 15.9 & 46.5 &&  27.6  &  65.3 && 8.1 & 33.9 \\
GaitRef \cite{zhu2023gaitref} && 17.7 & 50.2 &&  29.9  &  69.4  &&  9.5  & 37.2  \\
SMPLGait \cite{zheng2022gait} &&  18.8 & 51.9 &&  25.2  &  63.4  &&  14.6 &  44.3 \\
PSE (Ours) && 21.2 & 65.3 && 23.2 & 68.6 && 19.9 & 63.2  \\
\midrule
DME \cite{guo2022multi}&& 25.0 & 63.8  && 30.4 & 68.8 && 21.5 & 60.5 \\
PSTA \cite{wang2021pyramid} &&  33.6 & 67.3 && 32.1 & 66.0 && 34.5 & 68.1\\
CAL \cite{gu2022clothes} &&  34.9 & 71.4 && 34.7 & 71.0 && 35.0 & 71.7\\
TCL Net \cite{hou2020temporal} && 31.3 & 65.6 && 31.0 & 65.1 && 31.5 & 65.9\\
Attn-CL+rerank \cite{pathak2020video} &&  27.6 & 61.8 && 26.9 & 60.5 && 28.1 & 62.6\\
AAE (Ours) &&  38.3 & 81.8 &&  37.6 & 79.0  && 39.5 & 83.7 \\
\midrule
ShARc && \textbf{41.1} & \textbf{83.0}  &&  \textbf{39.4} & \textbf{80.7}  &&  \textbf{42.2} & \textbf{84.5} \\
\bottomrule
\end{tabular}
}
\caption{Identification results on BRIAR dataset. } 
\label{tab:oumvlp}
\end{table}

\textit{Appearance-based Recognition Model.} For input frames ${f_i}$, we first employ a ResNet-50 \cite{he2016deep} network which is pretrained on ImageNet \cite{russakovsky2015imagenet} dataset for feature encoding to get their $H\times W\times C$ feature maps $A_i$ before spatial pooling. For AAE, we follow \cite{wang2021pyramid} and use the patch level encoding for building a three-layer pyramid architecture with two different levels of attentions: temporal attention (TA) between two consecutive frames, and spatial attention (SA) of each frame. TA and SA of the same layer of the pyramid share weight, while those from different layers do not. The output attention is the same size as the input feature $A_i$, so we apply point-wise production for each input attention-feature pair and sum them up as the output, which is the input for the next level of the pyramid. For averaging aggregation, we set $\gamma$ as 0, which degrades the function to a binarized representation, following our results for ablation studies in Sec.~\ref{sec:expd}. After having the two features from AAE, we concatenate them to represent the appearance of the person.

\textit{Training and Inference.} Due to the network's complexity, we do not combine shape and appearance during training but train them individually end-to-end with their own inputs. For the shape-based network, we use the Adam optimizer for 180,000 iterations and set the initial learning rate as $1e^{-3}$. The learning rate is decayed to $\frac{1}{10}$ three times at iterations 30,000, 90,000, and 150,000. For the appearance-based method, we follow \cite{wang2021pyramid} and train the network for 500 epochs, using the Adam optimizer with an initial learning rate of $3.5e^{-4}$. We decay the learning rate by $0.3$ at steps 70, 140, 210, 310, and 410 during training.

During inference, we follow \cite{wieczorek2021unreasonable} by using centroid representation when registering the features of gallery examples via averaging all the features with the same ID. If there are multiple single frames, as gallery examples in BRIAR, we first combine the frames for the same ID as a `pseudo video' before sending it into the network for feature extraction. When querying an example with the gallery, we use the cosine distance to find the highest score in the gallery for shape score $S_{shape}(V)$ and Euclidean distance for appearance score $S_{app}(V)$ following existing gait recognition works \cite{chao2019gaitset}. If videos are shorter than 8 frames, we resample the frames until we have 8 frames for appearance feature extraction, and if the video is longer than 8 frames, we separate the video into several groups of 8 frames and average the results after extracting the features from all the groups. 

\textbf{Baseline Methods and Metrics.} In our experiment, we compare our method with some state-of-the-art person-reID methods on different datasets. For MEVID \cite{davila2023mevid}, we compared with CAL \cite{gu2022clothes}, AGRL \cite{wu2020adaptive}, BiCnet-TKS \cite{hou2021bicnet}, TCLNet \cite{hou2020temporal}, PSTA \cite{wang2021pyramid}, PiT \cite{zang2022multidirection}, STMN \cite{eom2021video}, Attn-CL \cite{pathak2020video}, Attn-CL+rerank \cite{pathak2020video}, and AP3D \cite{gu2020appearance} following the official results in MEVID \cite{davila2023mevid}. For CCVID \cite{gu2022clothes}, we compared with CAL \cite{gu2022clothes} following their original paper setting. For the comparison on BRIAR, we select some re-ID methods \cite{wang2021pyramid,gu2022clothes,hou2020temporal,pathak2020video} based on their performance on MEVID, as well as including some gait-based recognition methods \cite{chao2019gaitset,fan2020gaitpart,lin2021gaitgl,guo2022multi} for comparison.
For evaluation metrics, we use rank accuracies and mAP (mean average precision) for evaluation on these datasets.

\begin{table}
\begin{center}
\def\lw{1.6}
\def\ls{0.02}
\resizebox{\columnwidth}{!}
{
\begin{tabular}{p{3cm} >{\raggedleft\arraybackslash}p{\lw cm} >{\raggedleft\arraybackslash}p{\lw cm} >{\raggedleft\arraybackslash}p{\lw cm} >{\raggedleft\arraybackslash}p{\lw cm} >{\raggedleft\arraybackslash}p{\lw cm}}
\toprule
Methods & mAP & Rank-1 & Rank-5 & Rank-10 & Rank-20 \\
\midrule
BiCnet-TKS \cite{hou2021bicnet} & 6.3 & 19.0 & 35.1 & 40.5 & 52.9\\
PiT \cite{zang2022multidirection} & 13.6 & 34.2 & 55.4 & 63.3 & 70.6\\
STMN \cite{eom2021video} & 	11.3 & 31.0 & 54.4 & 65.5 & 72.5\\
AP3D \cite{gu2020appearance} & 15.9 & 39.6 & 56.0 & 63.3 & 76.3 \\
TCLNet \cite{hou2020temporal} & 23.0& 48.1 &60.1 & 69.0 &76.3	\\
PSTA \cite{wang2021pyramid} & 21.2 & 46.2 & 60.8 & 69.6 & 77.8\\
AGRL \cite{wu2020adaptive} & 19.1 & 48.4 & 62.7 & 70.6 & 77.9\\
Attn-CL \cite{pathak2020video} & 18.6 & 42.1 & 56.0 & 63.6 & 73.1\\
Attn-CL+rerank \cite{pathak2020video} & 25.9 & 46.5 & 59.8 & 64.6 & 71.8\\
CAL \cite{gu2022clothes} & 27.1 & 52.5 & 66.5 & 73.7 & 80.7 \\
\midrule
PSE & 10.6 & 25.9 & 39.9 & 48.7 & 62.7\\
AAE & \textbf{29.6} & 59.2 & \textbf{70.3} & \textbf{77.2} & \textbf{83.2}\\
ShARc & \textbf{29.6} & \textbf{59.5} & \textbf{70.3} & \textbf{77.2} & 82.9\\
\bottomrule
\end{tabular}}
\end{center}
\caption{Rank accuracy and mAP on MEVID dataset. Results for existing methods are from official MEVID \cite{davila2023mevid} implementation.}
\label{table:mevid}
\end{table}

\subsection{Results and Analysis}\label{sec:result} 
To compare with existing methods, we present the results for different baseline methods on the BRIAR, MEVID, and CCVID datasets in Tables \ref{tab:oumvlp}, \ref{table:mevid}, and \ref{table:ccvid}, respectively. In addition, we conduct some further ablation studies along with visualizations of the attention generated by the appearance branch for analysis of why the appearance model still works for clothes changing cases.

\textbf{Results for person identification.} As our main experiment, we have compared with all the three datasets with state-of-the-art methods in Table~\ref{tab:oumvlp}, \ref{table:mevid} and \ref{table:ccvid} respectively. Note that all these three datasets are describing the clothes change settings in the re-ID task, which is more complex than the existing person re-ID tasks with same outfit. We have the following observations.

\textbf{\textit{(i) Identification Performance.}} Our proposed method, ShARc, demonstrates significant performance improvements on all three datasets when compared to other state-of-the-art methods. For instance, on the BRIAR dataset, SHARc, after combining shape and appearance, achieves a 6.2\% and 11.6\% improvement in rank-1 and rank-20 accuracy, substantially outperforming other state-of-the-art methods. Moreover, on the other clothes-changing datasets, our method attains a 2.5\% and 7.5\% improvement in mAP and Rank-1 accuracy on MEVID \cite{davila2023mevid}, as well as a 4.6\% and 8.0\% improvement on CCVID \cite{gu2022clothes}. Note that we follow \cite{davila2023mevid} not using centroid averaging for gallery on MEVID. In addition, unlike BRIAR and CCVID, activities in MEVID do not include specific walking patterns, which results in a limited contribution from the PSE when combined with the appearance-based method, AAE.

Apart from the overall dataset results, we note that gait-based methods \cite{chao2019gaitset,fan2020gaitpart,lin2021gaitgl,guo2022multi} and appearance-based methods \cite{wang2021pyramid,gu2022clothes,hou2020temporal,pathak2020video} display different performance differences for the two types of activities, standing and walking. On the BRIAR dataset, gait-based methods \cite{chao2019gaitset,fan2020gaitpart,lin2021gaitgl} struggle with stationary sequences. Although DME \cite{guo2022multi}\footnote{The BRIAR dataset has included more subjects compared to the version used in DME, making it considerably more challenging.} demonstrates reasonable performance by incorporating masked RGB images into the gait branch, it still faces challenges when gait information is not available. In contrast, appearance-based methods exhibit slightly better performance with stationary videos compared to walking sequences, as stationary videos have less blurred boundaries due to reduced motion.

\begin{table}
\begin{center}
\def\lw{1.8}
\def\ls{0.02}
\resizebox{\columnwidth}{!}
{
\begin{tabular}{p{3cm}p{\lw cm}<{\centering}p{\lw cm}<{\centering}p{\ls cm}p{\lw cm}<{\centering}p{\lw cm}<{\centering}}
\toprule
\multirow{2}{*}{Method}&  \multicolumn{2}{c}{General} &&\multicolumn{2}{c}{CC}\\
\cline{2-3} \cline{5-6}   \\ [-8pt]
& Rank-1 & mAP && Rank-1 & mAP \\
\midrule
GaitNet \cite{song2019gaitnet} & 62.6 & 56.5 && 57.7 & 49.0\\
GaitSet \cite{chao2019gaitset} &  81.9 & 73.2 && 71.0 &62.1\\
PSE (Ours) & 83.9 & 86.5 &&  77.1 & 85.0\\
\midrule
CAL-baseline \cite{gu2022clothes} & 78.3 & 75.4 && 77.3  & 73.9\\
CAL Triplet \cite{gu2022clothes} & 81.5 & 78.1 && 81.1 & 77.0\\
CAL \cite{gu2022clothes} & 82.6 & 81.3 && 81.7 &79.6\\
AAE (Ours)  & 89.7 & 89.9 &&   84.6 & 84.8 \\
\midrule
ShARc & \textbf{89.8} & \textbf{90.2} && \textbf{84.7} & \textbf{85.2} \\ 
\bottomrule
\end{tabular}}
\end{center}
\caption{Rank-1 accuracy and mAP on CCVID dataset. CC includes the videos specifically for clothes changing, while general include both same and different clothing.}
\label{table:ccvid}
\end{table}

\textbf{\textit{(ii) Shape and Appearance Analysis.}} Apart from comparing our method with existing methods, we also separate the shape and appearance models, PSE and AAE, to evaluate their individual contributions in ShARc. We present the results in Table~\ref{tab:oumvlp}, \ref{table:mevid}, and \ref{table:ccvid}. Our appearance-based approach, AAE, demonstrates a substantial improvement over other appearance-based methods and achieves the best performance. This suggests that the averaging aggregation is indeed effective in providing supplementary information not captured by attention-based methods, thus helping to alleviate the overfitting problem. Furthermore, our shape-based model, PSE, not only outperforms other gait-based methods but also shows relatively robust performance on stationary videos, indicating that the integration of body shape features allows the model to better understand and distinguish between individuals, particularly when gait is unavailable.

It is worth noting that on datasets involving clothes-changing scenarios, such as BRIAR where the outfits between gallery and query videos are strictly different, appearance-based methods consistently outperform shape-based methods, even when both gait and body shape information are available. As shown in Table~\ref{tab:oumvlp}, appearance-based methods continue to surpass gait and body shape-based methods under different clothing conditions. One possible explanation for this observation is that the process of generating body shape (SMPL) and gait (silhouettes) features directly from RGB frames introduces noise or increases information loss during the preprocessing stage. This results in a degradation of the extracted features' quality and their effectiveness in the re-identification task. 

On the other hand, appearance-based methods can effectively leverage the rich information provided by RGB images to focus on relevant areas, even when the patterns of outfits differ between gallery and probe videos. This finding highlights the potential limitations of human-designed features, such as gait patterns or 3-D body shape, which despite being specifically and carefully designed for certain tasks, may still lead to information loss and underperform when compared to machine-designed features. In the final part of this section, we will present visualizations that further illustrate the effectiveness of our appearance-based method in handling clothes-changing scenarios.

\begin{table}[tb]
\centering
\def\lw{1.3}
\def\lk{2.4}
\def\ls{0.06}
\resizebox{\linewidth}{!}
{
\begin{tabular}{p{1.5cm}p{\ls cm}p{\lw cm}<{\centering}p{\lw cm}<{\centering}p{\lw cm}<{\centering}p{\lw cm}<{\centering}p{\lw cm}<{\centering}} \toprule

\multirow{1}{*}{Distances} && 200m & 400m  & 500m & 1000m & UAV  \\
\midrule
PSE   &&  38.5 & 38.2 & 35.7 &   5.3 & 25.9\\
AAE &&  60.6 & 56.3 & 51.2 &     10.5 & 30.7\\
ShARc   &&  64.3 & 60.4 & 56.0 & 10.5 & 36.4\\
\bottomrule
\end{tabular}
}
\caption{Rank-1 accuracy for different distances in BRIAR.} 
\label{tab:distance}
\end{table}

\begin{table}[tb]
\centering
\def\lw{1.3}
\def\lk{2.4}
\def\ls{0.06}
\resizebox{0.86\linewidth}{!}
{
\begin{tabular}{p{3cm}p{\ls cm}p{\lw cm}<{\centering}p{\lw cm}<{\centering}p{\lw cm}<{\centering}} \toprule

\multirow{1}{*}{Distances} && Rank 1   & Rank 5  & Rank 20 \\
\midrule
PSE && 21.2 & 44.9 & 65.3 \\
\ \ \ \ w/o binarized sil. && {8.7} & {20.7} & {40.1}  \\
\ \ \ \ w/o skeletons && 19.7 & 35.6 & 63.4\\
\ \ \ \ w/o 3-D shape && 8.7 & 20.1 & 37.6 \\
\midrule
AAE && 38.3 & 63.7 & 81.8 \\
\ \ \ \ w/o att. && 29.1 & 51.3 & 68.9 \\
\ \ \ \ w/o avg. && 33.0  & 57.2 & 77.5 \\
\ \ \ \ w/o centroid \cite{wieczorek2021unreasonable} && 30.9 & 56.1 & 75.4 \\
\bottomrule
\end{tabular}
}
\caption{Ablation results for different components in ShARc. `att' and `avg' are attention-based and averaging aggregations.} 
\label{tab:modal}
\end{table}

\textbf{Ablation results.} Since the BRIAR dataset provides valuable information, such as the exact distance at which images are captured and the impact of different types of activities in the sequences, we conduct ablation experiments on a sampled validation set derived from the training sequences to analyze the selection of weights when fusing the scores from the shape and appearance models.

\textit{Distances.} We present the performance of our method across various distances in Table~\ref{tab:distance}. We select five distance variations from the BRIAR dataset: 200 meters, 400 meters, 500 meters, 1000 meters, and video captured from UAV cameras. Generally, performance is better at shorter distances. However, we see a significant performance drop at 1000 meters, where the bodies in images are nearly indistinguishable. The results for UAV-captured images aren't as strong as those at 200 meters. This is due to the incomplete body images, as the UAV images are taken with the head occluding the whole body. The performance decline of PSE is less compared to AAE, showing its relative robustness in identification when occlusion is present.

\begin{table}[tb]
\centering
\def\lw{1}
\def\lk{2.4}
\def\ls{0.06}
\resizebox{0.8\linewidth}{!}
{
\begin{tabular}{p{2.5cm}p{\ls cm}p{\lw cm}<{\centering}p{\lw cm}<{\centering}p{\lw cm}<{\centering}p{\lw cm}<{\centering}} \toprule

\multirow{1}{*}{Gamma} && 1   & 0.2  & 0.1 & 0\\
\midrule
Rank 1  && 35.1 & 36.6 & 37.5 & 38.3 \\
\bottomrule
\end{tabular}
}
\caption{Rank-1 accuracy for feature flattening for AvA.} 
\label{tab:ff}
\end{table}

\begin{table}[tb]
\centering
\def\lw{1.2}
\def\lk{2.4}
\def\ls{0.06}
\resizebox{0.86\linewidth}{!}
{
\begin{tabular}{p{1.2cm}p{\ls cm}p{\lw cm}<{\centering}p{\lw cm}<{\centering}p{\lw cm}<{\centering}p{\lw cm}<{\centering}p{\lw cm}<{\centering}} \toprule

\multirow{1}{*}{App.} && 0.95   & 0.9  & 0.8 & 0.7 & 0.6 \\
\multirow{1}{*}{Shape} && 0.05   & 0.1  & 0.2 & 0.3 & 0.4 \\
\midrule
Rank 1 &&  91.1 & 91.4 & 91.0 & 90.2  & 88.4  \\
\bottomrule
\end{tabular}
}
\caption{Rank-1 accuracy for the selection of $\alpha$.} 
\label{tab:weight}
\end{table}

\textit{Model Components Ablations.} Our pipeline consists of multiple sub-modules, and we analyze the individual contribution of each component in both branches. For gait representation, we have two components: masked RGB and binarized silhouettes. We investigate the contributions of binarized silhouette masks and masked RGB images independently. It is important to note that the masked RGB images in this case are resized to a smaller scale, similar to binarized silhouettes, to provide information about the separation of body parts rather than directly using appearance for training. To remove each component in the network, we zero out the corresponding input for analysis.

We show the results in Table~\ref{tab:modal}. For the shape-based branch, masked RGB contributes the most, while 3-D body shape and binarized silhouettes contribute almost equally. Compared to other modalities, 3-D masked RGB images precisely provide more internal content for the gait branch, enabling the network to understand the boundary of different body parts and the movement of each part. For the appearance branch, we find that both aggregation contribute similarly to the final performance, and the combination of both yields the best results. Furthermore, using centroid averaging \cite{wieczorek2021unreasonable} when registering gallery examples also has a significant contribution to the final performance.

\textit{Feature Flattening.} For the flattening layer in averaging aggregation, we analyze the different Gamma and their corresponding results in Table~\ref{tab:ff}. When Gamma is 1, we have simple averaging across all the features. We observe that with higher gamma values, our performance improves, indicating that the results exhibit more discriminative patterns. When gamma is infinity, the final feature representation becomes binarized and yields the best performance.

\textit{Choice of $\alpha$.} To combine the two modalities, we construct a small validation set from the training data to analyze the weights between appearance and shape models, and present the results in Table~\ref{tab:weight}. We find that when the weight is 0.9 for appearance and 0.1 for shape, the model achieves the best performance. For shape-based methods, we use Euclidean distance instead of cosine distance; thus, 0.1 does not imply that it contributes minimally, but rather serves as a scaling factor for $S_{shape}$ during combination.

\begin{figure}
    \centering
    \includegraphics[width=.96\linewidth]{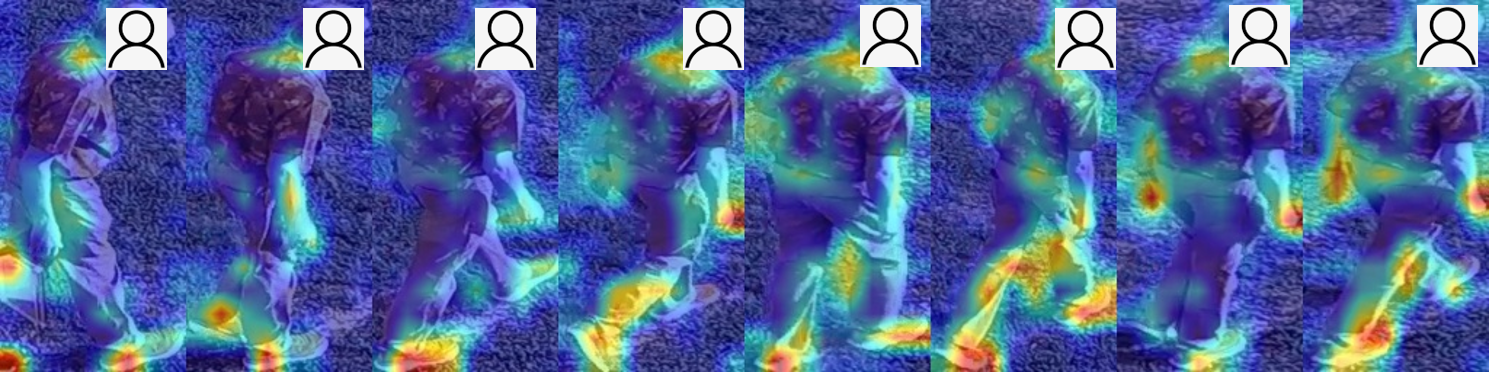}
    \includegraphics[width=.96\linewidth]{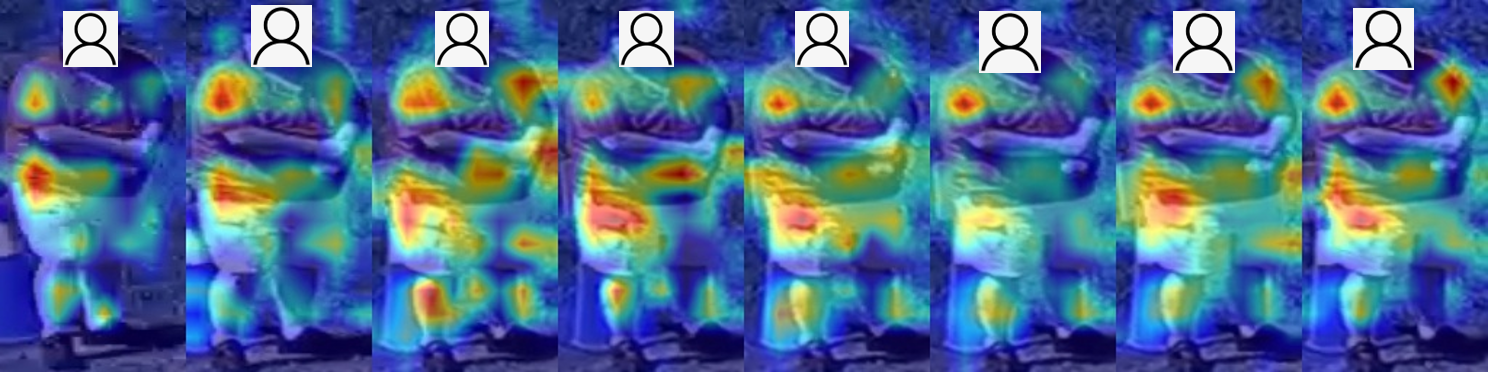}
    \caption{Attention generated from appearance model for (a) a walking sequence and (b) a stationary video for two examples taken from 100 meters distance category.}
    \label{fig:attnvis}
\end{figure}

\textbf{Visualization for Appearance Branch.} In the BRIAR dataset, where query and gallery images feature distinct outfits, we use GradCam~\cite{selvaraju2017grad} to visualize network focus. Figure~\ref{fig:attnvis} presents two examples taken from 100-meter-distance cameras, one during walking and another while stationary. For walking videos, the network focuses mainly on the lower body and arms, suggesting implicit pose pattern extraction. In stationary scenarios, attention is directed towards the waist and shoulders, important areas for discerning body shape. We observe this trend across multiple examples, although quantification has not been performed.

\section{Conclusion}
In this paper, we introduce ShARc, a shape and appearance-based method for identification in-the-wild. Our approach explicitly explores the contribution of body shape and appearance to the model with two encoders, pose and shape encoder for body shape and motion, and aggregated appearance encoder for human appearance. ShARc is able to handle most of the challenges for identification in the wild, such as occlusion, non-walking sequences, change of clothes, and image degradations. We have compared our method on three public datasets, including BRIAR, CCVID, and MEVID, and show state-of-the-art performance.

{\small
\bibliographystyle{ieee_fullname}
\bibliography{egbib}

\begin{thebibliography}{10}\itemsep=-1pt

\bibitem{an2020performance}
Weizhi An, Shiqi Yu, Yasushi Makihara, Xinhui Wu, Chi Xu, Yang Yu, Rijun Liao,
  and Yasushi Yagi.
\newblock Performance evaluation of model-based gait on multi-view very large
  population database with pose sequences.
\newblock {\em TBIOM}, 2(4):421--430, 2020.

\bibitem{chao2019gaitset}
Hanqing Chao, Yiwei He, Junping Zhang, and Jianfeng Feng.
\newblock Gaitset: Regarding gait as a set for cross-view gait recognition.
\newblock In {\em AAAI}, pages 8126--8133, 2019.

\bibitem{chen2017rethinking}
Liang-Chieh Chen, George Papandreou, Florian Schroff, and Hartwig Adam.
\newblock Rethinking atrous convolution for semantic image segmentation.
\newblock {\em arXiv preprint arXiv:1706.05587}, 2017.

\bibitem{cho2022part}
Yoonki Cho, Woo~Jae Kim, Seunghoon Hong, and Sung-Eui Yoon.
\newblock Part-based pseudo label refinement for unsupervised person
  re-identification.
\newblock In {\em CVPR}, pages 7308--7318, 2022.

\bibitem{cornett2023expanding}
David Cornett, Joel Brogan, Nell Barber, Deniz Aykac, Seth Baird, Nicholas
  Burchfield, Carl Dukes, Andrew Duncan, Regina Ferrell, Jim Goddard, et~al.
\newblock Expanding accurate person recognition to new altitudes and ranges:
  The briar dataset.
\newblock In {\em WACV}, pages 593--602, 2023.

\bibitem{dai2018video}
Ju Dai, Pingping Zhang, Dong Wang, Huchuan Lu, and Hongyu Wang.
\newblock Video person re-identification by temporal residual learning.
\newblock {\em TIP}, 28(3):1366--1377, 2018.

\bibitem{davila2023mevid}
Daniel Davila, Dawei Du, Bryon Lewis, Christopher Funk, Joseph Van~Pelt,
  Roderic Collins, Kellie Corona, Matt Brown, Scott McCloskey, Anthony Hoogs,
  et~al.
\newblock Mevid: Multi-view extended videos with identities for video person
  re-identification.
\newblock In {\em WACV}, pages 1634--1643, 2023.

\bibitem{deng2019arcface}
Jiankang Deng, Jia Guo, Niannan Xue, and Stefanos Zafeiriou.
\newblock Arcface: Additive angular margin loss for deep face recognition.
\newblock In {\em CVPR}, pages 4690--4699, 2019.

\bibitem{duan2019uniformface}
Yueqi Duan, Jiwen Lu, and Jie Zhou.
\newblock Uniformface: Learning deep equidistributed representation for face
  recognition.
\newblock In {\em CVPR}, pages 3415--3424, 2019.

\bibitem{eom2021video}
Chanho Eom, Geon Lee, Junghyup Lee, and Bumsub Ham.
\newblock Video-based person re-identification with spatial and temporal memory
  networks.
\newblock In {\em ICCV}, pages 12036--12045, 2021.

\bibitem{fan2022opengait}
Chao Fan, Junhao Liang, Chuanfu Shen, Saihui Hou, Yongzhen Huang, and Shiqi Yu.
\newblock Opengait: Revisiting gait recognition toward better practicality.
\newblock {\em arXiv preprint arXiv:2211.06597}, 2022.

\bibitem{fan2020gaitpart}
Chao Fan, Yunjie Peng, Chunshui Cao, Xu Liu, Saihui Hou, Jiannan Chi, Yongzhen
  Huang, Qing Li, and Zhiqiang He.
\newblock Gaitpart: Temporal part-based model for gait recognition.
\newblock In {\em CVPR}, pages 14225--14233, 2020.

\bibitem{fu2019sta}
Yang Fu, Xiaoyang Wang, Yunchao Wei, and Thomas Huang.
\newblock Sta: Spatial-temporal attention for large-scale video-based person
  re-identification.
\newblock In {\em AAAI}, volume~33, pages 8287--8294, 2019.

\bibitem{gao2018revisiting}
Jiyang Gao and Ram Nevatia.
\newblock Revisiting temporal modeling for video-based person reid.
\newblock {\em arXiv preprint arXiv:1805.02104}, 2018.

\bibitem{gao2022semantic}
Zan Gao, Hongwei Wei, Weili Guan, Jie Nie, Meng Wang, and Shenyong Chen.
\newblock A semantic-aware attention and visual shielding network for
  cloth-changing person re-identification.
\newblock {\em arXiv preprint arXiv:2207.08387}, 2022.

\bibitem{gu2022clothes}
Xinqian Gu, Hong Chang, Bingpeng Ma, Shutao Bai, Shiguang Shan, and Xilin Chen.
\newblock Clothes-changing person re-identification with rgb modality only.
\newblock In {\em CVPR}, pages 1060--1069, 2022.

\bibitem{gu2020appearance}
Xinqian Gu, Hong Chang, Bingpeng Ma, Hongkai Zhang, and Xilin Chen.
\newblock Appearance-preserving 3d convolution for video-based person
  re-identification.
\newblock In {\em ECCV}, pages 228--243. Springer, 2020.

\bibitem{guo2022multi}
Yuxiang Guo, Cheng Peng, Chun~Pong Lau, and Rama Chellappa.
\newblock Multi-modal human authentication using silhouettes, gait and rgb.
\newblock {\em arXiv preprint arXiv:2210.04050}, 2022.

\bibitem{he2016deep}
Kaiming He, Xiangyu Zhang, Shaoqing Ren, and Jian Sun.
\newblock Deep residual learning for image recognition.
\newblock In {\em CVPR}, pages 770--778, 2016.

\bibitem{he2021partial}
Tianyu He, Xu Shen, Jianqiang Huang, Zhibo Chen, and Xian-Sheng Hua.
\newblock Partial person re-identification with part-part correspondence
  learning.
\newblock In {\em CVPR}, pages 9105--9115, 2021.

\bibitem{hong2021fine}
Peixian Hong, Tao Wu, Ancong Wu, Xintong Han, and Wei-Shi Zheng.
\newblock Fine-grained shape-appearance mutual learning for cloth-changing
  person re-identification.
\newblock In {\em CVPR}, pages 10513--10522, 2021.

\bibitem{hou2021bicnet}
Ruibing Hou, Hong Chang, Bingpeng Ma, Rui Huang, and Shiguang Shan.
\newblock Bicnet-tks: Learning efficient spatial-temporal representation for
  video person re-identification.
\newblock In {\em CVPR}, pages 2014--2023, 2021.

\bibitem{hou2020temporal}
Ruibing Hou, Hong Chang, Bingpeng Ma, Shiguang Shan, and Xilin Chen.
\newblock Temporal complementary learning for video person re-identification.
\newblock In {\em ECCV}, pages 388--405. Springer, 2020.

\bibitem{hou2020gln}
Saihui Hou, Chunshui Cao, Xu Liu, and Yongzhen Huang.
\newblock Gait lateral network: Learning discriminative and compact
  representations for gait recognition.
\newblock In {\em ECCV}, pages 382--398, 2020.

\bibitem{ionescu2013human3}
Catalin Ionescu, Dragos Papava, Vlad Olaru, and Cristian Sminchisescu.
\newblock Human3. 6m: Large scale datasets and predictive methods for 3d human
  sensing in natural environments.
\newblock {\em TPAMI}, 36(7):1325--1339, 2013.

\bibitem{ji2020attention}
Zilong Ji, Xiaolong Zou, Xiaohan Lin, Xiao Liu, Tiejun Huang, and Si Wu.
\newblock An attention-driven two-stage clustering method for unsupervised
  person re-identification.
\newblock In {\em ECCV}, pages 20--36, 2020.

\bibitem{jin2022cloth}
Xin Jin, Tianyu He, Kecheng Zheng, Zhiheng Yin, Xu Shen, Zhen Huang, Ruoyu
  Feng, Jianqiang Huang, Zhibo Chen, and Xian-Sheng Hua.
\newblock Cloth-changing person re-identification from a single image with gait
  prediction and regularization.
\newblock In {\em CVPR}, pages 14278--14287, 2022.

\bibitem{kim2022adaface}
Minchul Kim, Anil~K Jain, and Xiaoming Liu.
\newblock Adaface: Quality adaptive margin for face recognition.
\newblock In {\em CVPR}, pages 18750--18759, 2022.

\bibitem{li2019global}
Jianing Li, Jingdong Wang, Qi Tian, Wen Gao, and Shiliang Zhang.
\newblock Global-local temporal representations for video person
  re-identification.
\newblock In {\em ICCV}, pages 3958--3967, 2019.

\bibitem{li2018diversity}
Shuang Li, Slawomir Bak, Peter Carr, and Xiaogang Wang.
\newblock Diversity regularized spatiotemporal attention for video-based person
  re-identification.
\newblock In {\em CVPR}, pages 369--378, 2018.

\bibitem{li2014deepreid}
Wei Li, Rui Zhao, Tong Xiao, and Xiaogang Wang.
\newblock Deepreid: Deep filter pairing neural network for person
  re-identification.
\newblock In {\em CVPR}, pages 152--159, 2014.

\bibitem{li2020end}
Xiang Li, Yasushi Makihara, Chi Xu, Yasushi Yagi, Shiqi Yu, and Mingwu Ren.
\newblock End-to-end model-based gait recognition.
\newblock In {\em ACCV}, 2020.

\bibitem{li2021diverse}
Yulin Li, Jianfeng He, Tianzhu Zhang, Xiang Liu, Yongdong Zhang, and Feng Wu.
\newblock Diverse part discovery: Occluded person re-identification with
  part-aware transformer.
\newblock In {\em CVPR}, pages 2898--2907, 2021.

\bibitem{liang2022gaitedge}
Junhao Liang, Chao Fan, Saihui Hou, Chuanfu Shen, Yongzhen Huang, and Shiqi Yu.
\newblock Gaitedge: Beyond plain end-to-end gait recognition for better
  practicality.
\newblock {\em arXiv preprint arXiv:2203.03972}, 2022.

\bibitem{lin2021gaitgl}
Beibei Lin, Shunli Zhang, and Xin Yu.
\newblock Gait recognition via effective global-local feature representation
  and local temporal aggregation.
\newblock In {\em ICCV}, pages 14648--14656, 2021.

\bibitem{lin2014microsoft}
Tsung-Yi Lin, Michael Maire, Serge Belongie, James Hays, Pietro Perona, Deva
  Ramanan, Piotr Doll{\'a}r, and C~Lawrence Zitnick.
\newblock Microsoft coco: Common objects in context.
\newblock In {\em ECCV}, pages 740--755, 2014.

\bibitem{liu2019spatially}
Chih-Ting Liu, Chih-Wei Wu, Yu-Chiang~Frank Wang, and Shao-Yi Chien.
\newblock Spatially and temporally efficient non-local attention network for
  video-based person re-identification.
\newblock {\em arXiv preprint arXiv:1908.01683}, 2019.

\bibitem{liu2017video}
Hao Liu, Zequn Jie, Karlekar Jayashree, Meibin Qi, Jianguo Jiang, Shuicheng
  Yan, and Jiashi Feng.
\newblock Video-based person re-identification with accumulative motion
  context.
\newblock {\em TCSVT}, 28(10):2788--2802, 2017.

\bibitem{liu2020disentangling}
Ziyu Liu, Hongwen Zhang, Zhenghao Chen, Zhiyong Wang, and Wanli Ouyang.
\newblock Disentangling and unifying graph convolutions for skeleton-based
  action recognition.
\newblock In {\em CVPR}, pages 143--152, 2020.

\bibitem{loper2015smpl}
Matthew Loper, Naureen Mahmood, Javier Romero, Gerard Pons-Moll, and Michael~J
  Black.
\newblock Smpl: A skinned multi-person linear model.
\newblock {\em TOG}, 34(6):1--16, 2015.

\bibitem{mclaughlin2016recurrent}
Niall McLaughlin, Jesus~Martinez Del~Rincon, and Paul Miller.
\newblock Recurrent convolutional network for video-based person
  re-identification.
\newblock In {\em CVPR}, pages 1325--1334, 2016.

\bibitem{mehta2017monocular}
Dushyant Mehta, Helge Rhodin, Dan Casas, Pascal Fua, Oleksandr Sotnychenko,
  Weipeng Xu, and Christian Theobalt.
\newblock Monocular 3d human pose estimation in the wild using improved cnn
  supervision.
\newblock In {\em 3DV}, pages 506--516, 2017.

\bibitem{nalty2022brief}
Chrisopher~B Nalty, Neehar Peri, Joshua Gleason, Carlos~D Castillo, Shuowen Hu,
  Thirimachos Bourlai, and Rama Chellappa.
\newblock A brief survey on person recognition at a distance.
\newblock {\em arXiv preprint arXiv:2212.08969}, 2022.

\bibitem{pathak2020video}
Priyank Pathak, Amir~Erfan Eshratifar, and Michael Gormish.
\newblock Video person re-id: Fantastic techniques and where to find them
  (student abstract).
\newblock In {\em AAAI}, volume~34, pages 13893--13894, 2020.

\bibitem{russakovsky2015imagenet}
Olga Russakovsky, Jia Deng, Hao Su, Jonathan Krause, Sanjeev Satheesh, Sean Ma,
  Zhiheng Huang, Andrej Karpathy, Aditya Khosla, Michael Bernstein, et~al.
\newblock Imagenet large scale visual recognition challenge.
\newblock {\em International journal of computer vision}, 115:211--252, 2015.

\bibitem{schroff2015facenet}
Florian Schroff, Dmitry Kalenichenko, and James Philbin.
\newblock Facenet: A unified embedding for face recognition and clustering.
\newblock In {\em CVPR}, pages 815--823, 2015.

\bibitem{selvaraju2017grad}
Ramprasaath~R Selvaraju, Michael Cogswell, Abhishek Das, Ramakrishna Vedantam,
  Devi Parikh, and Dhruv Batra.
\newblock Grad-cam: Visual explanations from deep networks via gradient-based
  localization.
\newblock In {\em ICCV}, pages 618--626, 2017.

\bibitem{shen2022lidar}
Chuanfu Shen, Chao Fan, Wei Wu, Rui Wang, George~Q Huang, and Shiqi Yu.
\newblock Lidar gait: Benchmarking 3d gait recognition with point clouds.
\newblock {\em arXiv preprint arXiv:2211.10598}, 2022.

\bibitem{si2018dual}
Jianlou Si, Honggang Zhang, Chun-Guang Li, Jason Kuen, Xiangfei Kong, Alex~C
  Kot, and Gang Wang.
\newblock Dual attention matching network for context-aware feature sequence
  based person re-identification.
\newblock In {\em CVPR}, pages 5363--5372, 2018.

\bibitem{song2019gaitnet}
Chunfeng Song, Yongzhen Huang, Yan Huang, Ning Jia, and Liang Wang.
\newblock Gaitnet: An end-to-end network for gait based human identification.
\newblock {\em PR}, 96:106988, 2019.

\bibitem{subramaniam2019co}
Arulkumar Subramaniam, Athira Nambiar, and Anurag Mittal.
\newblock Co-segmentation inspired attention networks for video-based person
  re-identification.
\newblock In {\em ICCV}, pages 562--572, 2019.

\bibitem{sun2019deep}
Ke Sun, Bin Xiao, Dong Liu, and Jingdong Wang.
\newblock Deep high-resolution representation learning for human pose
  estimation.
\newblock In {\em CVPR}, pages 5693--5703, 2019.

\bibitem{sun2021monocular}
Yu Sun, Qian Bao, Wu Liu, Yili Fu, Michael~J Black, and Tao Mei.
\newblock Monocular, one-stage, regression of multiple 3d people.
\newblock In {\em ICCV}, pages 11179--11188, 2021.

\bibitem{sun2018beyond}
Yifan Sun, Liang Zheng, Yi Yang, Qi Tian, and Shengjin Wang.
\newblock Beyond part models: Person retrieval with refined part pooling (and a
  strong convolutional baseline).
\newblock In {\em ECCV}, pages 480--496, 2018.

\bibitem{takemura2018multi}
Noriko Takemura, Yasushi Makihara, Daigo Muramatsu, Tomio Echigo, and Yasushi
  Yagi.
\newblock Multi-view large population gait dataset and its performance
  evaluation for cross-view gait recognition.
\newblock {\em TCVA}, 10(1):1--14, 2018.

\bibitem{teepe2021gaitgraph}
Torben Teepe, Ali Khan, Johannes Gilg, Fabian Herzog, Stefan H{\"o}rmann, and
  Gerhard Rigoll.
\newblock Gaitgraph: Graph convolutional network for skeleton-based gait
  recognition.
\newblock In {\em ICIP}, pages 2314--2318, 2021.

\bibitem{wang2020deep}
Jingdong Wang, Ke Sun, Tianheng Cheng, Borui Jiang, Chaorui Deng, Yang Zhao,
  Dong Liu, Yadong Mu, Mingkui Tan, Xinggang Wang, et~al.
\newblock Deep high-resolution representation learning for visual recognition.
\newblock {\em TPAMI}, 43(10):3349--3364, 2020.

\bibitem{wang2021pyramid}
Yingquan Wang, Pingping Zhang, Shang Gao, Xia Geng, Hu Lu, and Dong Wang.
\newblock Pyramid spatial-temporal aggregation for video-based person
  re-identification.
\newblock In {\em ICCV}, pages 12026--12035, 2021.

\bibitem{wen2016discriminative}
Yandong Wen, Kaipeng Zhang, Zhifeng Li, and Yu Qiao.
\newblock A discriminative feature learning approach for deep face recognition.
\newblock In {\em ECCV}, pages 499--515, 2016.

\bibitem{wieczorek2021unreasonable}
Miko{\l}aj Wieczorek, Barbara Rychalska, and Jacek D{\k{a}}browski.
\newblock On the unreasonable effectiveness of centroids in image retrieval.
\newblock In {\em ICONIP}, pages 212--223, 2021.

\bibitem{wu2020adaptive}
Yiming Wu, Omar El~Farouk Bourahla, Xi Li, Fei Wu, Qi Tian, and Xue Zhou.
\newblock Adaptive graph representation learning for video person
  re-identification.
\newblock {\em TIP}, 29:8821--8830, 2020.

\bibitem{xu2017jointly}
Shuangjie Xu, Yu Cheng, Kang Gu, Yang Yang, Shiyu Chang, and Pan Zhou.
\newblock Jointly attentive spatial-temporal pooling networks for video-based
  person re-identification.
\newblock In {\em ICCV}, pages 4733--4742, 2017.

\bibitem{yu2006framework}
Shiqi Yu, Daoliang Tan, and Tieniu Tan.
\newblock A framework for evaluating the effect of view angle, clothing and
  carrying condition on gait recognition.
\newblock In {\em ICPR}, volume~4, pages 441--444, 2006.

\bibitem{zang2022multidirection}
Xianghao Zang, Ge Li, and Wei Gao.
\newblock Multidirection and multiscale pyramid in transformer for video-based
  pedestrian retrieval.
\newblock {\em TII}, 18(12):8776--8785, 2022.

\bibitem{zhao2017deeply}
Liming Zhao, Xi Li, Yueting Zhuang, and Jingdong Wang.
\newblock Deeply-learned part-aligned representations for person
  re-identification.
\newblock In {\em ICCV}, pages 3219--3228, 2017.

\bibitem{zheng2022gait}
Jinkai Zheng, Xinchen Liu, Wu Liu, Lingxiao He, Chenggang Yan, and Tao Mei.
\newblock Gait recognition in the wild with dense 3d representations and a
  benchmark.
\newblock In {\em CVPR}, pages 20228--20237, 2022.

\bibitem{zheng2016mars}
Liang Zheng, Zhi Bie, Yifan Sun, Jingdong Wang, Chi Su, Shengjin Wang, and Qi
  Tian.
\newblock Mars: A video benchmark for large-scale person re-identification.
\newblock In {\em ECCV}, pages 868--884, 2016.

\bibitem{zheng2015scalable}
Liang Zheng, Liyue Shen, Lu Tian, Shengjin Wang, Jingdong Wang, and Qi Tian.
\newblock Scalable person re-identification: A benchmark.
\newblock In {\em ICCV}, pages 1116--1124, 2015.

\bibitem{zheng2017person}
Liang Zheng, Hengheng Zhang, Shaoyan Sun, Manmohan Chandraker, Yi Yang, and Qi
  Tian.
\newblock Person re-identification in the wild.
\newblock In {\em CVPR}, pages 1367--1376, 2017.

\bibitem{zhou2017see}
Zhen Zhou, Yan Huang, Wei Wang, Liang Wang, and Tieniu Tan.
\newblock See the forest for the trees: Joint spatial and temporal recurrent
  neural networks for video-based person re-identification.
\newblock In {\em CVPR}, pages 4747--4756, 2017.

\bibitem{zhu2022open}
Haidong Zhu, Ye Yuan, Yiheng Zhu, Xiao Yang, and Ram Nevatia.
\newblock Open: Order-preserving pointcloud encoder decoder network for body
  shape refinement.
\newblock In {\em ICPR}, pages 521--527, 2022.

\bibitem{zhu2023gaitref}
Haidong Zhu, Wanrong Zheng, Zhaoheng Zheng, and Ram Nevatia.
\newblock Gaitref: Gait recognition with refined sequential skeletons.
\newblock {\em arXiv preprint arXiv:2304.07916}, 2023.

\bibitem{zhu2023gait}
Haidong Zhu, Zhaoheng Zheng, and Ram Nevatia.
\newblock Gait recognition using 3-d human body shape inference.
\newblock In {\em WACV}, pages 909--918, 2023.

\bibitem{zhu2022pass}
Kuan Zhu, Haiyun Guo, Tianyi Yan, Yousong Zhu, Jinqiao Wang, and Ming Tang.
\newblock Pass: Part-aware self-supervised pre-training for person
  re-identification.
\newblock In {\em ECCV}, pages 198--214, 2022.

\end{thebibliography}
}

\end{document}